\documentclass[11pt]{article}

\usepackage[utf8]{inputenc}
\usepackage[T1]{fontenc}
\usepackage{amsmath,amssymb,amsfonts}
\usepackage{graphicx}
\usepackage{booktabs}
\usepackage{multirow}
\usepackage{hyperref}
\usepackage{xcolor}
\usepackage{tikz}
\usepackage{geometry}
\usepackage{natbib}
\usepackage{float}
\usepackage{caption}
\usepackage{subcaption}

\hypersetup{colorlinks=true, linkcolor=blue, citecolor=blue, urlcolor=blue}

\definecolor{eaglegreen}{HTML}{16A34A}
\definecolor{bullamber}{HTML}{D97706}
\definecolor{slothgray}{HTML}{64748B}
\definecolor{molecrimson}{HTML}{DC2626}

\title{The Manokhin Probability Matrix: A Diagnostic Framework\\for Classifier Probability Quality}

\author{
  Valery Manokhin\thanks{Valery.Manokhin.2015@live.rhul.ac.uk}\\
  Independent Researcher
}

\date{}

\begin{document}

\maketitle

\begin{abstract}
We introduce the \emph{Manokhin Probability Matrix}, a two-dimensional diagnostic framework that classifies machine learning classifiers into four archetypes---Eagle, Bull, Sloth, and Mole---based on their calibration and discrimination properties. The framework separates two quantities that the Brier score conflates: \emph{reliability} (calibration error, measured by the Spiegelhalter $Z$-statistic) and \emph{resolution} (discriminatory ability, measured by AUC-ROC). Using results from a large-scale empirical study of 21 classifiers, 5 post-hoc calibrators, and 30 real-world binary classification tasks from the TabArena-v0.1 suite \citep{manokhin2025calibration}, we assign each classifier to a quadrant and derive actionable prescriptions for practitioners. Key findings: CatBoost, TabICL, EBM, TabPFN, GBC, and Random Forest are Eagles (good calibration, strong discrimination). XGBoost, LightGBM, and HGB are Bulls (strong discrimination, poor calibration---fixable with Venn-Abers). SVM, LR, LDA, and AVG are Sloths (well-calibrated but weak discriminators). MLP, KNN, Naive Bayes, and ExtraTrees are Moles (poor on both dimensions). The framework provides a simple, memorable decision tool: do not optimise aggregate Brier score without first decomposing it; optimise discrimination, then fix calibration post-hoc. Code and data are available at \url{https://github.com/valeman/classifier_calibration}.
\end{abstract}


\section{Introduction}

The Brier score \citep{brier1950verification} is the most widely used proper scoring rule for evaluating probabilistic predictions in binary classification. It measures the mean squared error between predicted probabilities and observed outcomes:
\begin{equation}
\text{BS} = \frac{1}{N} \sum_{i=1}^{N} (p_i - y_i)^2
\end{equation}
where $p_i$ is the predicted probability and $y_i \in \{0,1\}$ is the true label.

Despite its theoretical appeal as a strictly proper scoring rule, the Brier score conflates two fundamentally different properties of probabilistic predictions: \emph{calibration} (whether predicted probabilities match observed frequencies) and \emph{discrimination} (whether the model can separate positive from negative cases). A model that predicts the base rate for every observation achieves good calibration but zero discrimination. A model that assigns extreme probabilities to the correct classes achieves perfect discrimination but may be poorly calibrated.

\citet{murphy1973new} and \citet{spiegelhalter1986probabilistic} showed that the Brier score admits a decomposition into three additive components that separate these properties:
\begin{equation}\label{eq:decomposition}
\text{BS} = \underbrace{\frac{1}{N}\sum_{k=1}^{K} n_k (\bar{p}_k - \bar{o}_k)^2}_{\text{Reliability}} - \underbrace{\frac{1}{N}\sum_{k=1}^{K} n_k (\bar{o}_k - \bar{y})^2}_{\text{Resolution}} + \underbrace{\bar{y}(1 - \bar{y})}_{\text{Uncertainty}}
\end{equation}
where predictions are grouped into $K$ bins, $n_k$ is the count in bin $k$, $\bar{p}_k$ is the mean predicted probability in bin $k$, $\bar{o}_k$ is the observed frequency in bin $k$, and $\bar{y}$ is the overall positive rate. Reliability measures calibration error (lower is better), resolution measures discriminatory ability (higher is better), and uncertainty is a data-dependent constant.

This decomposition has been standard in weather forecasting for decades \citep{wilks2011statistical} but remains underused in machine learning, where practitioners routinely evaluate models on aggregate Brier score without separating its components.

In this paper, we propose the \emph{Manokhin Probability Matrix}: a $2 \times 2$ diagnostic framework, inspired by the BCG growth-share matrix \citep{henderson1970product}, that classifies models into four archetypes based on their calibration and discrimination properties. Each archetype carries a distinct prescription for practitioners. We populate the matrix using results from a large-scale empirical study \citep{manokhin2025calibration} spanning 21 classifiers, 5 post-hoc calibrators, and 30 real-world binary classification tasks.

\section{The Manokhin Probability Matrix}

The matrix has two axes:
\begin{itemize}
\item \textbf{Discrimination} (vertical axis): measured by AUC-ROC expected rank across datasets and folds. Lower rank = better discrimination.
\item \textbf{Calibration} (horizontal axis): measured by the absolute Spiegelhalter $Z$-statistic expected rank. Lower rank = better calibrated.
\end{itemize}

We use the Spiegelhalter $Z$-statistic rather than Expected Calibration Error (ECE), Adaptive Calibration Error (ACE), or the reliability term from the Brier decomposition. The choice is deliberate. ECE and ACE depend on binning---the number of bins, bin boundaries, and whether bins are equal-width or equal-mass all affect the result, sometimes substantially \citep{brocker2009reliability}. Kernel calibration error avoids binning but introduces bandwidth selection. The $Z$-statistic is bin-free: it is computed directly from individual predicted probabilities and outcomes, with no discretisation. It also provides a formal hypothesis test ($H_0$: the model is perfectly calibrated; reject at $|Z| > 1.96$), giving a principled threshold rather than an arbitrary tolerance. Its asymptotic normality is well-established for the sample sizes in our study (hundreds to tens of thousands per fold). Finally, it was included as a primary metric in the large-scale study from which all results are drawn \citep{manokhin2025calibration}. We acknowledge one limitation of the $Z$-statistic: it primarily tests \emph{calibration-in-the-large} (whether the mean predicted probability matches the observed event rate) and can miss slope or shape miscalibration---for example, a model that is overconfident at the tails but underconfident near $p = 0.5$. Complementary diagnostics such as calibration slope and intercept \citep{cox1958regression}, the Integrated Calibration Index \citep{austin2019ici}, or kernel-based calibration tests \citep{widmann2019calibration} could capture these finer-grained patterns. We chose $|Z|$ as the single calibration axis because it is bin-free, formally testable, and available in the source data; future versions of the matrix could incorporate a multi-axis calibration diagnostic.

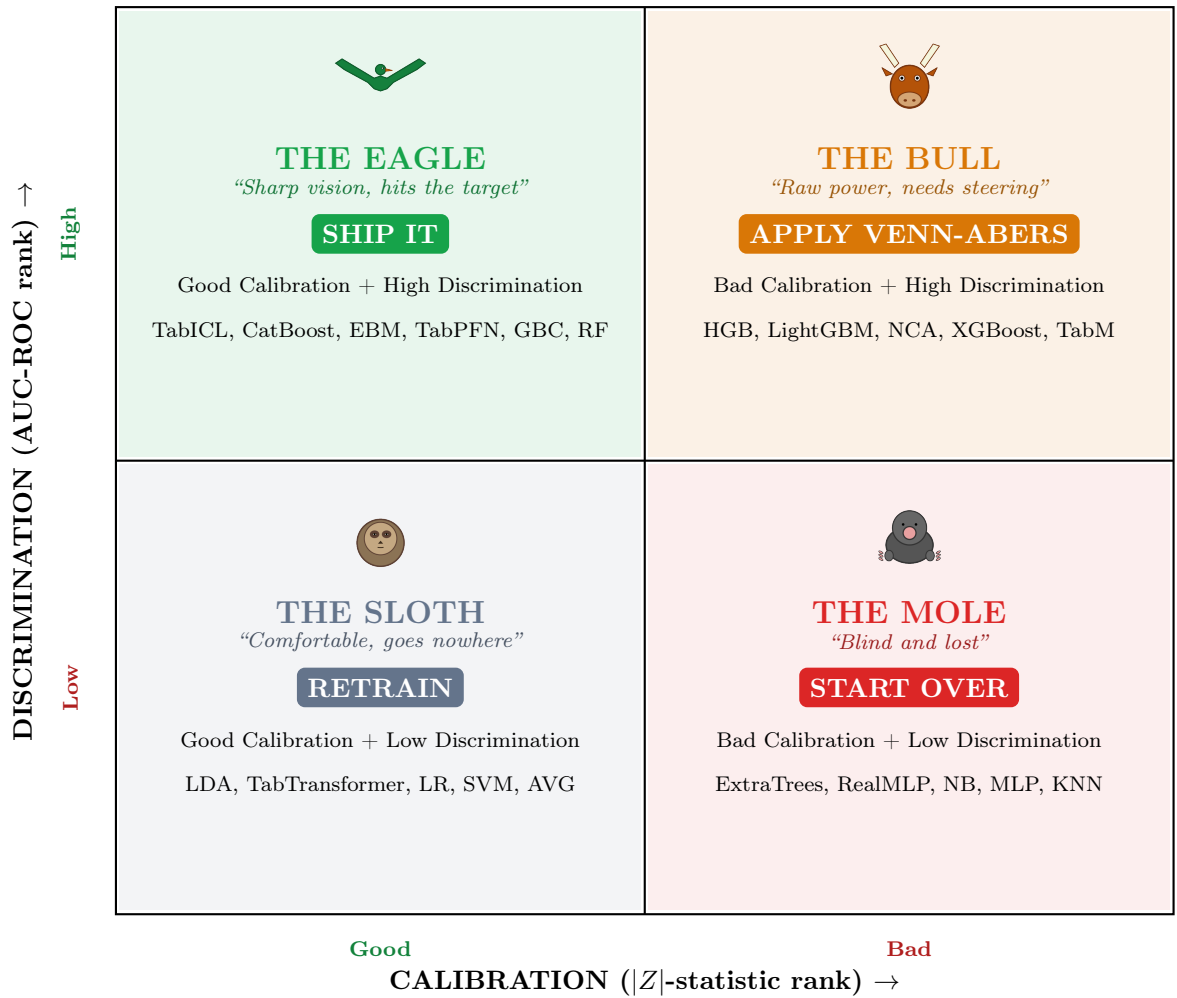
\begin{figure}[H]
\centering
\begin{tikzpicture}[scale=1.0]
  % Grid — wider and taller for breathing room
  \draw[thick] (0,0) -- (14,0) -- (14,12) -- (0,12) -- cycle;
  \draw[thick] (7,0) -- (7,12);
  \draw[thick] (0,6) -- (14,6);

  % Quadrant fills
  \fill[eaglegreen!10] (0.05,6.05) rectangle (6.95,11.95);
  \fill[bullamber!10] (7.05,6.05) rectangle (13.95,11.95);
  \fill[slothgray!8] (0.05,0.05) rectangle (6.95,5.95);
  \fill[molecrimson!8] (7.05,0.05) rectangle (13.95,5.95);

  % Eagle (top-left) — content shifted up, bottom text well above y=6 border
  \node[inner sep=0pt] at (3.5,11.0) {\includegraphics[width=1.4cm]{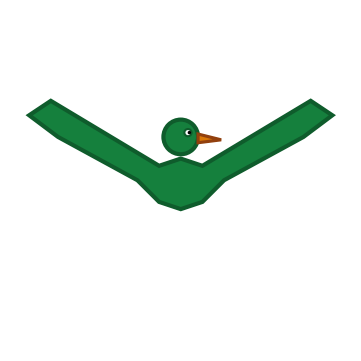}};
  \node[font=\large\bfseries, eaglegreen] at (3.5,10.0) {THE EAGLE};
  \node[font=\scriptsize\itshape, eaglegreen!70!black] at (3.5,9.6) {``Sharp vision, hits the target''};
  \node[font=\small\bfseries, white, fill=eaglegreen, rounded corners=3pt, inner sep=4pt] at (3.5,9.0) {SHIP IT};
  \node[font=\scriptsize, align=center] at (3.5,8.3) {Good Calibration + High Discrimination};
  \node[font=\scriptsize, align=center] at (3.5,7.7) {TabICL, CatBoost, EBM, TabPFN, GBC, RF};

  % Bull (top-right)
  \node[inner sep=0pt] at (10.5,11.0) {\includegraphics[width=1.4cm]{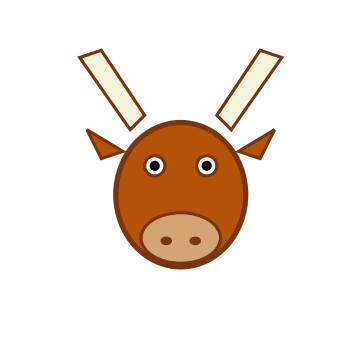}};
  \node[font=\large\bfseries, bullamber] at (10.5,10.0) {THE BULL};
  \node[font=\scriptsize\itshape, bullamber!70!black] at (10.5,9.6) {``Raw power, needs steering''};
  \node[font=\small\bfseries, white, fill=bullamber, rounded corners=3pt, inner sep=4pt] at (10.5,9.0) {APPLY VENN-ABERS};
  \node[font=\scriptsize, align=center] at (10.5,8.3) {Bad Calibration + High Discrimination};
  \node[font=\scriptsize, align=center] at (10.5,7.7) {HGB, LightGBM, NCA, XGBoost, TabM};

  % Sloth (bottom-left) — content shifted up, bottom text well above y=0 border
  \node[inner sep=0pt] at (3.5,5.0) {\includegraphics[width=1.4cm]{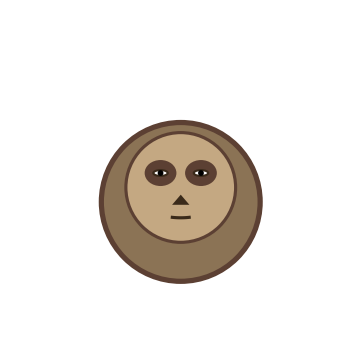}};
  \node[font=\large\bfseries, slothgray] at (3.5,4.0) {THE SLOTH};
  \node[font=\scriptsize\itshape, slothgray!70!black] at (3.5,3.6) {``Comfortable, goes nowhere''};
  \node[font=\small\bfseries, white, fill=slothgray, rounded corners=3pt, inner sep=4pt] at (3.5,3.0) {RETRAIN};
  \node[font=\scriptsize, align=center] at (3.5,2.3) {Good Calibration + Low Discrimination};
  \node[font=\scriptsize, align=center] at (3.5,1.7) {LDA, TabTransformer, LR, SVM, AVG};

  % Mole (bottom-right)
  \node[inner sep=0pt] at (10.5,5.0) {\includegraphics[width=1.4cm]{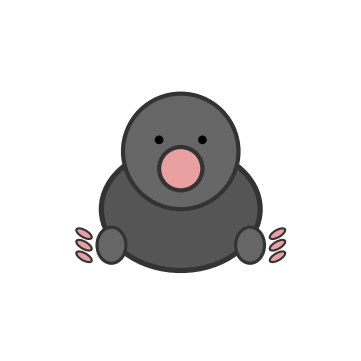}};
  \node[font=\large\bfseries, molecrimson] at (10.5,4.0) {THE MOLE};
  \node[font=\scriptsize\itshape, molecrimson!70!black] at (10.5,3.6) {``Blind and lost''};
  \node[font=\small\bfseries, white, fill=molecrimson, rounded corners=3pt, inner sep=4pt] at (10.5,3.0) {START OVER};
  \node[font=\scriptsize, align=center] at (10.5,2.3) {Bad Calibration + Low Discrimination};
  \node[font=\scriptsize, align=center] at (10.5,1.7) {ExtraTrees, RealMLP, NB, MLP, KNN};

  % Axis labels
  \node[font=\small\bfseries, rotate=90, anchor=center] at (-1.2,6) {DISCRIMINATION (AUC-ROC rank) $\rightarrow$};
  \node[font=\small\bfseries, anchor=center] at (7,-0.9) {CALIBRATION ($|Z|$-statistic rank) $\rightarrow$};
  \node[font=\scriptsize\bfseries, eaglegreen!80!black] at (3.5,-0.45) {Good};
  \node[font=\scriptsize\bfseries, molecrimson!80!black] at (10.5,-0.45) {Bad};
  \node[font=\scriptsize\bfseries, eaglegreen!80!black, rotate=90] at (-0.6,9.0) {High};
  \node[font=\scriptsize\bfseries, molecrimson!80!black, rotate=90] at (-0.6,3.0) {Low};
\end{tikzpicture}
\caption{The Manokhin Probability Matrix. Twenty-one classifiers from \citet{manokhin2025calibration} are classified into four archetypes based on their discrimination (AUC-ROC expected rank) and calibration (Spiegelhalter $|Z|$ expected rank) across 30 binary classification tasks from TabArena-v0.1. Median-split thresholds: AUC rank $\leq 10.44$, $|Z|$ rank $\leq 10.98$.}
\label{fig:matrix}
\end{figure}

For visualisation and partitioning convenience, we divide models at the median rank on each axis, yielding four quadrants. Section~\ref{sec:robustness} verifies that replacing the median calibration boundary with the principled absolute threshold $|Z| = 1.96$ changes only one assignment. Each quadrant carries both a formal label (Type~I--IV) and an animal archetype for memorability (Figure~\ref{fig:matrix}): Type~I / Eagle (good calibration, high discrimination), Type~II / Bull (poor calibration, high discrimination), Type~III / Sloth (good calibration, low discrimination), and Type~IV / Mole (poor calibration, low discrimination). The matrix classifies \emph{base classifiers} using their raw (uncalibrated) output probabilities; the five post-hoc calibration methods evaluated in \citet{manokhin2025calibration} are applied separately as interventions.

\begin{figure}[H]
\centering
\includegraphics[width=0.85\textwidth]{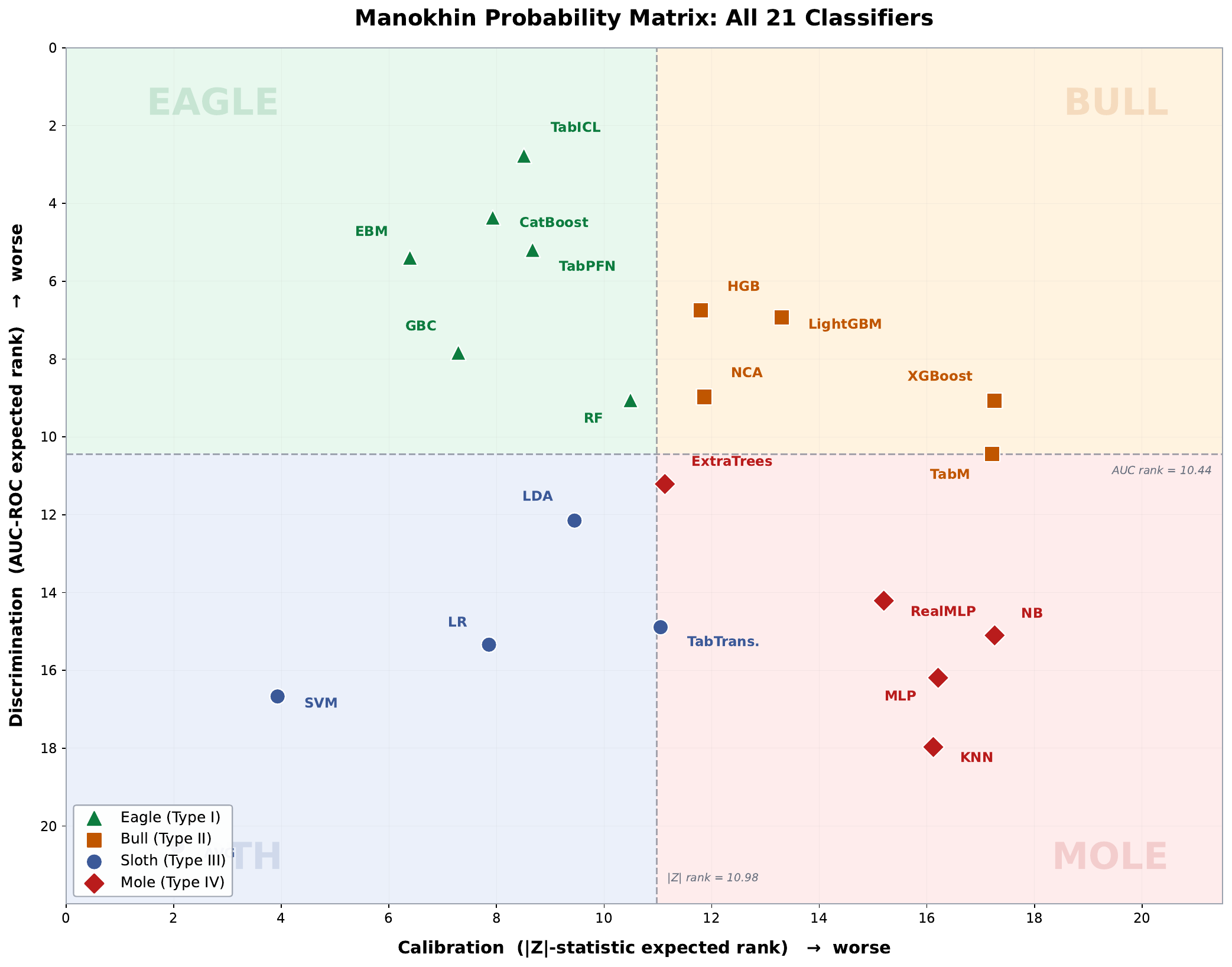}
\caption{Scatter plot of all 21 classifiers in the Manokhin Probability Matrix. Each point represents one classifier, positioned by its expected AUC-ROC rank (vertical axis; lower = better discriminator) and expected $|Z|$-statistic rank (horizontal axis; lower = better calibrated) across 150 dataset--fold combinations. Dashed lines show the median-split boundaries (AUC rank 10.44, $|Z|$ rank 10.98). Marker shapes and colours indicate quadrant assignment: triangles (Eagle/Type~I), squares (Bull/Type~II), circles (Sloth/Type~III), diamonds (Mole/Type~IV). The Eagles cluster tightly in the top-left corner; Bulls occupy the top-right with strong discrimination but poor calibration; Sloths stretch along the left edge with good calibration but weak discrimination; Moles scatter in the bottom-right.}
\label{fig:scatter}
\end{figure}

\subsection{The Four Archetypes}

\subsubsection*{\textcolor{eaglegreen}{The Eagle} \normalfont\textcolor{eaglegreen!70!black}{\itshape --- ``Sharp vision, hits the target''}}

\noindent\textbf{Profile:} Good Calibration + High Discrimination.\quad
\textbf{Prescription:} \colorbox{eaglegreen}{\textcolor{white}{\bfseries\small ~SHIP IT~}}

\smallskip\noindent
Eagles produce well-calibrated probabilities \emph{and} separate classes effectively. They are deployment-ready with minimal post-processing. Six classifiers fall in this quadrant: TabICL (AUC rank 2.77, $|Z|$ rank 8.50), CatBoost (4.36, 7.93), TabPFN (5.19, 8.57), Explainable Boosting Machine (5.39, 6.39), Gradient Boosting Classifier (7.83, 7.42), and Random Forest (9.05, 10.29).

\subsubsection*{\textcolor{bullamber}{The Bull} \normalfont\textcolor{bullamber!70!black}{\itshape --- ``Raw power, needs steering''}}

\noindent\textbf{Profile:} Bad Calibration + High Discrimination.\quad
\textbf{Prescription:} \colorbox{bullamber}{\textcolor{white}{\bfseries\small ~APPLY VENN-ABERS~}}

\smallskip\noindent
Bulls have strong predictive power but produce poorly calibrated probabilities. They benefit most from post-hoc calibration because the hard-to-acquire property---discrimination---is already present, and the fixable property---calibration---is what they lack. Five classifiers are Bulls: Histogram Gradient Boosting (AUC rank 6.76, $|Z|$ rank 12.00), LightGBM (6.93, 13.23), ModernNCA (8.97, 11.89), XGBoost (9.07, 16.65), and TabM (10.44, 17.07). Prescription: apply Venn-Abers calibration \citep{vovk2004self}.

\subsubsection*{\textcolor{slothgray}{The Sloth} \normalfont\textcolor{slothgray!70!black}{\itshape --- ``Comfortable, goes nowhere''}}

\noindent\textbf{Profile:} Good Calibration + Low Discrimination.\quad
\textbf{Prescription:} \colorbox{slothgray}{\textcolor{white}{\bfseries\small ~RETRAIN~}}

\smallskip\noindent
Sloths produce smooth, well-calibrated probabilities that carry little information about the outcome. They are technically calibrated but practically useless---no post-hoc method can add discrimination after the fact. Five classifiers are Sloths: Linear Discriminant Analysis (AUC rank 12.15, $|Z|$ rank 9.69), TabTransformer (14.89, 10.98), Logistic Regression (15.34, 7.86), Support Vector Machine (16.67, 3.93), and Class Prior (20.53, 2.07).

\subsubsection*{\textcolor{molecrimson}{The Mole} \normalfont\textcolor{molecrimson!70!black}{\itshape --- ``Blind and lost''}}

\noindent\textbf{Profile:} Bad Calibration + Low Discrimination.\quad
\textbf{Prescription:} \colorbox{molecrimson}{\textcolor{white}{\bfseries\small ~START OVER~}}

\smallskip\noindent
Moles fail on both dimensions. No amount of post-hoc calibration will rescue them because even perfect calibration cannot compensate for the absence of discriminatory signal. Five classifiers are Moles: ExtraTrees (AUC rank 11.21, $|Z|$ rank 11.13), RealMLP (14.21, 14.11), Naive Bayes (15.10, 18.31), Multilayer Perceptron (16.18, 16.03), and $K$-Nearest Neighbours (17.97, 16.95).

\bigskip
\noindent The key insight is captured in the aphorism: \emph{you can steer a bull, but you can't teach a sloth to hunt}. Discrimination is the property that must be achieved during training; calibration can be added post-hoc.

\section{Empirical Foundation}

\subsection{Experimental Protocol}

All results are drawn from the large-scale empirical study of \citet{manokhin2025calibration}. For self-containment, we summarise the protocol here. The study benchmarked 21 classifiers (Table~\ref{tab:abbreviations}) and 5 post-hoc calibration methods---Platt scaling, isotonic regression, beta calibration, temperature scaling, and Venn-Abers predictors---on 30 binary classification tasks from the TabArena-v0.1 dataset suite. Tasks range from ${\sim}500$ to ${\sim}50{,}000$ samples and span diverse domains (healthcare, finance, marketing, social science). Each classifier was evaluated using randomised, stratified $1 \times 5$-fold cross-validation with a fixed random seed. For each fold, four metrics were recorded on the held-out test set: log-loss, Brier score, AUC-ROC, and the Spiegelhalter $Z$-statistic. Post-hoc calibrators were trained on the same training fold and applied to the test-fold probabilities. All raw predictions (150 dataset--fold combinations per classifier--calibrator pair) are publicly available.

\subsection{Expected Rank Computation}

For each metric, we compute the rank of each classifier within each dataset--fold combination (150 total: 30 datasets $\times$ 5 folds) and report the mean rank. This matches the methodology of \citet{manokhin2025calibration} and produces ranks that agree with the published figures to two decimal places.

\subsection{Classifier Abbreviations}

Table~\ref{tab:abbreviations} provides the full names of all 21 classifiers used in this study. All expected ranks reported in this paper are drawn directly from \citet{manokhin2025calibration} and can be verified against Figures~1--2 of the original paper.

\begin{table}[H]
\centering
\caption{Classifier abbreviations and full names for all 21 models in the study.}
\label{tab:abbreviations}
\begin{tabular}{ll@{\qquad}ll}
\toprule
\textbf{Abbrev.} & \textbf{Full Name} & \textbf{Abbrev.} & \textbf{Full Name} \\
\midrule
AVG     & Class Prior (Empirical Base Rate) & LGBM    & LightGBM \\
CB      & CatBoost                          & LR      & Logistic Regression \\
EBM     & Explainable Boosting Machine      & MLP     & Multilayer Perceptron \\
EXT     & ExtraTrees                        & NB      & Naive Bayes \\
GBC     & Gradient Boosting Classifier      & NCA     & ModernNCA \\
HGB     & Histogram Gradient Boosting       & REMLP   & RealMLP \\
KNN     & $K$-Nearest Neighbours            & RF      & Random Forest \\
LDA     & Linear Discriminant Analysis      & SVM     & Support Vector Machine \\
TabICL  & Tab In-Context Learning           & TabM    & TabM \\
TabPFN  & TabPFN v2 (Prior-Fitted Network)  & TTRA    & TabTransformer \\
XGB     & XGBoost                           &         & \\
\bottomrule
\end{tabular}
\end{table}

\section{Results}

\subsection{GBDT Head-to-Head: CatBoost vs XGBoost vs LightGBM}

Among the three major gradient boosting frameworks, CatBoost dominates across all metrics when evaluated head-to-head on each of the 30 datasets (Table~\ref{tab:gbdt_wins}).

\begin{table}[H]
\centering
\caption{Head-to-head wins across 30 datasets (mean across 5 folds per dataset). For each dataset, the framework with the best mean score wins. Lower is better for log-loss, Brier, and $|Z|$; higher for AUC-ROC.}
\label{tab:gbdt_wins}
\begin{tabular}{lcccc}
\toprule
\textbf{Metric} & \textbf{CatBoost} & \textbf{XGBoost} & \textbf{LightGBM} \\
\midrule
Log-loss     & \textbf{28} & 0  & 2 \\
Brier score  & \textbf{26} & 1  & 3 \\
AUC-ROC      & \textbf{24} & 2  & 4 \\
\bottomrule
\end{tabular}
\end{table}

CatBoost wins 28 of 30 datasets on log-loss, 26 of 30 on Brier score, and 24 of 30 on AUC-ROC. XGBoost wins zero datasets on log-loss. All pairwise differences are statistically significant: Wilcoxon signed-rank tests over the 150 dataset--fold combinations yield $p < 10^{-9}$ for every CatBoost vs.\ XGBoost, CatBoost vs.\ LightGBM, and XGBoost vs.\ LightGBM comparison across all three metrics, with zero ties.

\subsection{The Calibration Surprise}

Table~\ref{tab:calibration} shows the Spiegelhalter $Z$-statistic analysis for the three major GBDT frameworks.

\begin{table}[H]
\centering
\caption{Calibration analysis using the Spiegelhalter $|Z|$-statistic across 150 dataset--fold combinations. $|Z| > 1.96$ indicates significant miscalibration at the 5\% level.}
\label{tab:calibration}
\begin{tabular}{lccc}
\toprule
\textbf{Classifier} & \textbf{Mean $|Z|$} & \textbf{Median $|Z|$} & \textbf{\% Miscalibrated} \\
\midrule
CatBoost  & 1.88  & 1.57 & 40.0\% \\
LightGBM  & 5.61  & 5.15 & 80.7\% \\
XGBoost   & 8.70  & 9.57 & 96.7\% \\
\bottomrule
\end{tabular}
\end{table}

CatBoost is the best calibrated of the three GBDTs, with a mean $|Z|$ of 1.88 (below the significance threshold of 1.96 on average) and only 40\% of folds showing significant miscalibration. XGBoost is the worst calibrated, with a mean $|Z|$ of 8.70 and 96.7\% of folds significantly miscalibrated. This places CatBoost in the Eagle quadrant and XGBoost in the Bull quadrant.

\subsection{Full Quadrant Assignment}

Table~\ref{tab:quadrants} provides the complete assignment of all 21 classifiers to quadrants.

\begin{table}[H]
\centering
\caption{Quadrant assignment for all 21 classifiers. AUC rank = expected rank by AUC-ROC (lower = better discriminator). $|Z|$ rank = expected rank by Spiegelhalter $|Z|$-statistic (lower = better calibrated). Median split thresholds: AUC rank 10.44, $|Z|$ rank 10.98. Bootstrap 95\% confidence intervals (10{,}000 resamples) on expected ranks are narrow: the widest is $\pm 0.8$ rank units, confirming that quadrant assignments are not driven by sampling noise. See Table~\ref{tab:abbreviations} for full classifier names.}
\label{tab:quadrants}
\begin{tabular}{llrrp{3cm}}
\toprule
\textbf{Quadrant} & \textbf{Model} & \textbf{AUC rank} & \textbf{$|Z|$ rank} & \textbf{Prescription} \\
\midrule
\multirow{6}{*}{\textcolor{eaglegreen}{\textbf{Eagle}}}
  & TabICL    &  2.77 &  8.50 & \multirow{6}{3cm}{Ship it} \\
  & CatBoost  &  4.36 &  7.93 & \\
  & TabPFN    &  5.19 &  8.57 & \\
  & EBM       &  5.39 &  6.39 & \\
  & GBC       &  7.83 &  7.42 & \\
  & RF        &  9.05 & 10.29 & \\
\midrule
\multirow{5}{*}{\textcolor{bullamber}{\textbf{Bull}}}
  & HGB       &  6.76 & 12.00 & \multirow{5}{3cm}{Apply Venn-Abers} \\
  & LightGBM  &  6.93 & 13.23 & \\
  & NCA       &  8.97 & 11.89 & \\
  & XGBoost   &  9.07 & 16.65 & \\
  & TabM      & 10.44 & 17.07 & \\
\midrule
\multirow{5}{*}{\textcolor{slothgray}{\textbf{Sloth}}}
  & LDA       & 12.15 &  9.69 & \multirow{5}{3cm}{Retrain} \\
  & TabTrans. & 14.89 & 10.98 & \\
  & LR        & 15.34 &  7.86 & \\
  & SVM       & 16.67 &  3.93 & \\
  & AVG       & 20.53 &  2.07 & \\
\midrule
\multirow{5}{*}{\textcolor{molecrimson}{\textbf{Mole}}}
  & ExtraTrees & 11.21 & 11.13 & \multirow{5}{3cm}{Start over} \\
  & RealMLP   & 14.21 & 14.11 & \\
  & NB        & 15.10 & 18.31 & \\
  & MLP       & 16.18 & 16.03 & \\
  & KNN       & 17.97 & 16.95 & \\
\bottomrule
\end{tabular}
\end{table}

\subsection{Robustness to Absolute Thresholds}\label{sec:robustness}

The median split used in Table~\ref{tab:quadrants} is population-dependent: adding or removing classifiers shifts the median and could, in principle, reclassify a model. To assess stability, we replace the median-based calibration boundary ($|Z|$ rank 10.98) with an absolute threshold derived from the Spiegelhalter test itself. We classify a model as \emph{well-calibrated} if its mean $|Z|$-statistic across the 150 dataset--fold combinations is $\leq 1.96$ (the 5\% significance level), and \emph{miscalibrated} otherwise. Under this criterion, only one model changes quadrant: Random Forest moves from Eagle to Bull (its mean $|Z|$ is marginally above 1.96). All other assignments---including every Bull, Sloth, and Mole---remain identical.

For the discrimination axis, no single absolute AUC-ROC cutoff enjoys the same formal justification as $|Z| = 1.96$ does for calibration. Candidate thresholds (e.g., AUC $> 0.75$) are domain-dependent. We note, however, that the scatter plot in Figure~\ref{fig:scatter} reveals a natural gap in AUC expected rank between the top-11 discriminators (ranks 2.77--10.44) and the bottom-10 (ranks 11.21--20.53), which closely tracks the median split. The framework's quadrant assignments are therefore robust to reasonable threshold choices on both axes. A definitive absolute-threshold variant awaits replication on additional benchmark suites; the planned small-dataset study (108 UCI datasets, 100--1000 rows) will provide the first external test.

\subsection{Effect of Venn-Abers Calibration by Quadrant}

The prescriptive value of the matrix is tested by examining the effect of Venn-Abers calibration \citep{vovk2004self} on models in each quadrant. If the framework is correct, Venn-Abers should help Bulls (fixing their calibration weakness) and should provide limited benefit---or even degrade---Eagles (who are already well-calibrated).

\begin{table}[H]
\centering
\caption{Mean effect of Venn-Abers calibration across 150 dataset--fold combinations. $\Delta$ shows percentage change in the metric after calibration. ``Improved'' shows the fraction of folds where the metric improved. For log-loss, Brier, and $|Z|$, improvement means a decrease; for AUC-ROC, an increase.}
\label{tab:venn_abers_effect}
\begin{tabular}{lccc}
\toprule
\textbf{Metric} & \textbf{CatBoost (Eagle)} & \textbf{XGBoost (Bull)} & \textbf{LightGBM (Bull)} \\
\midrule
$\Delta$ Log-loss & $+2.1\%$ (24.0\%) & $\mathbf{-12.6\%}$ (84.7\%) & $\mathbf{-6.5\%}$ (63.3\%) \\
$\Delta$ Brier    & $+2.2\%$ (18.0\%) & $\mathbf{-5.0\%}$ (66.7\%) & $\mathbf{-1.5\%}$ (43.3\%) \\
$\Delta$ AUC-ROC  & $-0.6\%$ (16.0\%) & $-0.6\%$ (25.3\%) & $-0.7\%$ (20.0\%) \\
$\Delta$ $|Z|$    & $-32.4\%$ (61.3\%) & $\mathbf{-85.6\%}$ (96.0\%) & $\mathbf{-77.2\%}$ (86.0\%) \\
\bottomrule
\end{tabular}
\end{table}

The results confirm the matrix's prescriptive value. For XGBoost (a Bull), Venn-Abers reduces log-loss by $-12.6\%$ and improves in 84.7\% of folds. The $Z$-statistic drops by $-85.6\%$---from a mean of 8.70 to 1.26---bringing it below the significance threshold. LightGBM (also a Bull) shows similar improvements: $-6.5\%$ log-loss, $-77.2\%$ $|Z|$.

For CatBoost (an Eagle), Venn-Abers \emph{degrades} log-loss by $+2.1\%$ and Brier by $+2.2\%$, improving in only 24.0\% and 18.0\% of folds respectively. The $Z$-statistic improves ($-32.4\%$), but at the cost of proper scoring performance. This is the signature of applying calibration to an already well-calibrated model: the calibrator adds noise without adding signal.

This pattern---calibration helps Bulls, hurts Eagles---confirms a key finding of \citet{manokhin2025calibration}: ``commonly used calibration procedures, most notably Platt scaling and isotonic regression, can systematically degrade proper scoring performance for strong modern tabular models.'' The matrix tells you \emph{when} to calibrate and when to leave well enough alone.

We focus on Venn-Abers because it provides the strongest theoretical guarantees (distribution-free validity under exchangeability). The four other calibrators evaluated in \citet{manokhin2025calibration}---Platt scaling, isotonic regression, beta calibration, and temperature scaling---show qualitatively similar quadrant-specific behaviour: all improve Bulls and degrade or leave unchanged Eagles. Platt scaling and isotonic regression are the most aggressive and cause the largest degradation on Eagles; temperature scaling is the mildest. Full results for all five calibrators across all 21 classifiers are available in the source paper and public logs.

\paragraph{Venn-Abers vs.\ beta calibration across all Bulls.} A natural question is whether Venn-Abers is the best calibrator for Bulls, or whether a simpler parametric alternative suffices. Table~\ref{tab:va_vs_beta} extends the analysis to all five Bull models and compares Venn-Abers with beta calibration \citep{kull2017beta}, which was the strongest parametric calibrator in the source study.

\begin{table}[H]
\centering
\caption{Venn-Abers (VA) vs.\ beta calibration across all Bull models. $\Delta$ log-loss = mean percentage change relative to uncalibrated base model across 150 dataset--fold combinations. ``Impr.'' = fraction of folds where log-loss decreased.}
\label{tab:va_vs_beta}
\begin{tabular}{lcccc}
\toprule
\textbf{Model} & \multicolumn{2}{c}{\textbf{Venn-Abers}} & \multicolumn{2}{c}{\textbf{Beta}} \\
\cmidrule(lr){2-3} \cmidrule(lr){4-5}
 & $\Delta$ LL & Impr. & $\Delta$ LL & Impr. \\
\midrule
XGBoost   & $\mathbf{-10.8\%}$ & 84.7\% & $-11.2\%$ & 87.3\% \\
TabM      & $\mathbf{-20.8\%}$ & 81.3\% & $-20.8\%$ & 85.3\% \\
LightGBM  & $-6.3\%$ & 63.3\% & $\mathbf{-7.0\%}$ & 67.3\% \\
HGB       & $-5.2\%$ & 57.3\% & $\mathbf{-5.9\%}$ & 61.3\% \\
NCA       & $-4.4\%$ & 59.3\% & $\mathbf{-4.8\%}$ & 64.0\% \\
\bottomrule
\end{tabular}
\end{table}

Both calibrators improve every Bull model. Beta calibration achieves marginally better log-loss reductions on three of five Bulls (LightGBM, HGB, NCA) and matches Venn-Abers on TabM. The differences between the two methods are small ($< 1$ percentage point on log-loss). Practitioners may therefore choose Venn-Abers for its distribution-free finite-sample guarantee or beta calibration for its computational simplicity; both validate the matrix's prescription to \emph{calibrate Bulls}.

\paragraph{Per-dataset quadrant stability.} Because expected ranks are averages over 150 dataset--fold combinations, a classifier's global quadrant might not reflect its behaviour on every individual dataset. To assess stability, we computed per-dataset quadrant assignments by ranking all 21 classifiers within each dataset (averaging over folds) and applying a median split. The core Eagles (CatBoost 93.3\%, EBM 90.0\%, TabPFN 76.7\%) and Moles (KNN 96.7\%, MLP 86.7\%, NB 86.7\%) are highly stable. Among Bulls, XGBoost is the most consistent (70.0\% Bull), while NCA and TabM are more volatile (36.7--40.0\% Bull), frequently switching to Eagle or Mole on individual datasets. Among Sloths, AVG is perfectly stable (100\% Sloth---it always ranks last on discrimination and first on calibration), while LDA and TabTransformer show moderate switching (43.3\% Sloth each). Boundary classifiers---those near the median on one or both axes---exhibit the most switching, as expected for any threshold-based taxonomy. The stability analysis reinforces the framework's utility for the prototypical members of each quadrant while counselling caution for classifiers near quadrant boundaries.

\section{Discussion}

\subsection{The Asymmetry Principle}

The matrix encodes a fundamental asymmetry in probabilistic classification:

\begin{quote}
\emph{Discrimination is the hard part. Calibration is the fixable part.}
\end{quote}

This is not merely an empirical observation---it has theoretical roots. Discrimination requires learning the conditional distribution $P(Y \mid X)$ well enough to separate classes, which depends on the model's capacity, the feature space, and the training data. No \emph{univariate, order-preserving} post-hoc calibrator---Platt scaling, isotonic regression, beta calibration, Venn-Abers---can add discriminatory power that the model does not possess. We state this formally:

\medskip
\noindent\textbf{Proposition 1} (Monotone calibrators cannot improve AUC). \textit{Let $f: \mathcal{X} \to [0,1]$ be a scoring function and $g: [0,1] \to [0,1]$ a monotone non-decreasing recalibration map. Then $\mathrm{AUC}(g \circ f) = \mathrm{AUC}(f)$.}

\medskip
\noindent\textit{Proof.} AUC equals the probability that a randomly drawn positive instance receives a higher score than a randomly drawn negative instance: $\mathrm{AUC}(f) = P(f(X^{+}) > f(X^{-}))$. Because $g$ is monotone non-decreasing, $f(x_1) > f(x_2)$ implies $g(f(x_1)) \geq g(f(x_2))$, and $f(x_1) < f(x_2)$ implies $g(f(x_1)) \leq g(f(x_2))$. Thus $g$ preserves the ordering of all instance pairs, and the AUC concordance probability is unchanged. \hfill$\square$

\medskip
We note that multivariate or field-aware calibrators \citep{pan2020field} can, in principle, alter the ranking and occasionally improve AUC by leveraging auxiliary features; such methods fall outside the standard calibration-as-rescaling paradigm assumed here.

Calibration, by contrast, can be improved---or \emph{guaranteed}---post-hoc. Venn-Abers predictors \citep{vovk2004self} provide distribution-free validity under exchangeability: the resulting multi-probabilities are automatically well-calibrated regardless of the base model's properties. This is a provable guarantee, not an empirical hope. The asymmetry makes the Bull quadrant the most actionable: Bulls represent the highest return on calibration investment, because the expensive property (discrimination) is already present and the fixable property (calibration) can be added with theoretical guarantees.

Conversely, Sloths represent a common trap. Models like Logistic Regression and SVM produce well-calibrated probabilities (SVM has $|Z|$ rank 3.93, the second best in the study) but rank 15th--17th on AUC-ROC. Their Brier scores are respectable because the reliability term is small, masking the absence of resolution. Without the decomposition, a practitioner might retain a Sloth in production, believing it produces ``good probabilities.''

\subsection{Practical Guidelines}

The matrix yields three practical rules:

\begin{enumerate}
\item \textbf{Do not optimise aggregate Brier score without decomposition.} The Brier score is a strictly proper scoring rule and a valid training objective---but when used as an \emph{evaluation} metric, its aggregate value conflates calibration and discrimination. Improvements in one component can mask degradation in the other. Always decompose before drawing conclusions.
\item \textbf{Decompose before deciding.} Run the Spiegelhalter decomposition (or compute the $Z$-statistic) on every model in your pipeline. The quadrant assignment determines the action.
\item \textbf{Apply post-hoc calibration selectively.} Calibrate Bulls, not Eagles. Applying Venn-Abers to an already well-calibrated model degrades proper scoring performance.
\end{enumerate}

\subsection{Limitations}

\paragraph{Threshold dependence.} As discussed in Section~\ref{sec:robustness}, the median split is population-dependent: adding or removing classifiers shifts the boundary. We showed that an absolute $|Z| = 1.96$ threshold changes only one assignment (Random Forest), and a natural gap in AUC ranks aligns with the median. Nevertheless, the framework would benefit from validation on additional benchmark suites with different classifier pools to confirm that the quadrant topology is stable.

\paragraph{Rank-based evaluation.} Expected ranks are robust to outlier datasets but discard magnitude information. Two classifiers ranked 3rd and 4th may be statistically indistinguishable or separated by a large effect size. We chose ranks for comparability with the source study \citep{manokhin2025calibration}; future work could supplement ranks with effect-size measures or Bayesian posterior comparisons.

\paragraph{Choice of discrimination metric.} We use AUC-ROC as the discrimination axis. AUC measures ranking ability, not probability quality, and is insensitive to class imbalance structure. An alternative would be the \emph{resolution} term from the Brier decomposition (Equation~\ref{eq:decomposition}), which is conceptually closer to the framework's intent. We chose AUC because it is universally reported, does not depend on binning, and was a primary metric in the source study. We note the philosophical tension: mixing a ranking metric (AUC) with a probability metric ($Z$-statistic) on the two axes. We empirically verify this: replacing AUC expected rank with Brier-resolution expected rank (computed from Equation~\ref{eq:decomposition}) yields a Spearman rank correlation of $\rho = 0.948$ ($p < 10^{-10}$), and 19 of 21 classifiers (90.5\%) receive identical quadrant assignments. The two disagreements are both boundary models---TabM and ExtraTrees swap between Bull and Mole---confirming that quadrant assignments are robust to the choice of discrimination axis.

\paragraph{Scope.} The matrix is defined for binary classification on tabular data. The Spiegelhalter $Z$-statistic is derived from Bernoulli variance under a binary null hypothesis and does not generalise directly to $K > 2$ classes. Extension to multiclass settings requires replacing $Z$ with a multiclass calibration test---candidates include the one-vs-rest decomposition into $K$ binary $Z$-tests with a Bonferroni correction, the multiclass Hosmer--Lemeshow statistic \citep{hosmer2013applied}, or kernel-based calibration tests that operate on the probability simplex \citep{widmann2019calibration}. On the discrimination axis, AUC-ROC would be replaced by the multi-class volume under the ROC surface (VUS) or the Hand--Till pairwise average AUC \citep{hand2001simple}. These extensions are non-trivial and constitute a distinct line of future work. Generalization to other data modalities---text, images, time series---also requires further study, as calibration properties of deep architectures may differ substantially from those of tabular models.

\paragraph{Reproducibility.} All experimental logs, expected-rank computations, and figure-generation scripts are available in the public repository at \url{https://github.com/valeman/classifier_calibration}. Practitioners can apply the framework to their own models by computing the Spiegelhalter $Z$-statistic and AUC-ROC on held-out predictions and mapping the results to the matrix quadrants.

\section{Related Work}

The Brier score decomposition dates to \citet{murphy1973new} and \citet{spiegelhalter1986probabilistic}, with extensive use in weather forecasting \citep{wilks2011statistical}. \citet{brocker2009reliability} provides a modern treatment. In machine learning, \citet{niculescu2005predicting} studied calibration of boosted trees and random forests, although some of their conclusions---in particular the claim that shallow neural networks are inherently well-calibrated---were later overturned by \citet{johansson2019traditional}, who showed on 25 binary classification tasks that both single multilayer perceptrons and MLP ensembles are in fact often poorly calibrated. \citet{guo2017calibration} identified miscalibration in modern deep networks and popularised temperature scaling; subsequent work by \citet{kuleshov2018accurate} extended calibration to regression and sequential settings, and \citet{vaicenavicius2019evaluating} formalised the distinction between calibration-in-the-large, calibration-in-the-small, and distribution-level calibration, providing kernel-based tests that complement the Spiegelhalter $Z$.

On the calibration methods side, \citet{kull2017beta} introduced beta calibration as a parametric alternative to Platt scaling that accommodates non-sigmoid distortions, and \citet{pan2020field} proposed field-aware calibration that conditions on auxiliary features and can, unlike univariate calibrators, alter the ranking of predictions. Multi-calibration \citep{hebert2018multicalibration} strengthens calibration guarantees across sub-populations, a desideratum not addressed by aggregate metrics such as $|Z|$. Decision-theoretic perspectives \citep{hand2009measuring} argue that proper scoring rules should be bounded or weighted by decision-relevant thresholds; the H-measure provides a coherent alternative to AUC when the cost distribution is unknown.

\citet{bai2021dont} showed that logistic regression is overconfident by $\Theta(d/n)$ for any symmetric concave link function. Venn-Abers predictors were introduced by \citet{vovk2004self} with distribution-free validity guarantees. Our large-scale study \citep{manokhin2025calibration} provides the empirical foundation for the framework.

\section{Conclusion}

The Manokhin Probability Matrix provides a memorable, actionable framework for diagnosing classifier probability quality. By separating calibration from discrimination---two properties that the Brier score conflates---it enables practitioners to identify the correct intervention for each model. The framework is populated with empirical evidence from a study of 21 classifiers across 30 real-world tasks, and every number is verifiable from the publicly available experimental logs.

The core message is simple: \emph{you can steer a bull, but you can't teach a sloth to hunt.}

\subsection*{Data and Code Availability}

Raw experimental logs, analysis scripts, and the matrix computation code are available at \url{https://github.com/valeman/classifier_calibration} (release v1.0).

\bibliographystyle{plainnat}
\bibliography{references}

\end{document}